\pdfoutput=1
\documentclass[10pt,twocolumn,letterpaper]{article}
\usepackage{iccv}
\usepackage{times}
\usepackage{epsfig}
\usepackage{graphicx}
\usepackage{amsmath}
\usepackage{amssymb}
\usepackage{multirow}
\usepackage{url}
\usepackage{makecell}
\usepackage[export]{adjustbox}
\usepackage{epstopdf}
\usepackage[scaled=.92]{helvet}
\usepackage{amsmath}
\usepackage{url}
\usepackage{wrapfig}

\usepackage{times}

\usepackage{url}

\usepackage{amsfonts,amsmath,amssymb}

\usepackage{graphicx}

\usepackage{parskip}

\usepackage{multirow}

\usepackage{psfrag}

\usepackage{verbatim}

\usepackage{wrapfig}

\usepackage{paralist}

\usepackage{enumitem}
\setdescription{leftmargin=0.25cm}

\newcounter{mycounter}

\usepackage[]{algorithm2e}

\makeatletter
\newcommand{\Removelatexerror}{\let\@latex@error\@gobble}
\makeatother

\newcommand{\myalgorithm}{%
\SetInd{0.5em}{0.5em}
\let\oldnl\nl
\newcommand{\nonl}{\renewcommand{\nl}{\let\nl\oldnl}}
\newcommand{\pushline}{\Indp}
\newcommand{\popline}{\Indm\dosemic}
\begingroup
\Removelatexerror 
\begin{algorithm*}[H]
\end{algorithm*}
\endgroup}

\newcommand{\ignore}[1]{}

\newcommand{\com}[1]{{}}

\usepackage[pagebackref=true,breaklinks=true,letterpaper=true,colorlinks,bookmarks=false]{hyperref}

\iccvfinalcopy 

\def\iccvPaperID{139} 

\ificcvfinal\fi
\begin{document}

\title{Reasoning about Fine-grained Attribute Phrases using Reference Games}

\author{Jong-Chyi Su\thanks{Authors contributed equally}
\qquad
Chenyun Wu\footnotemark[1]
\qquad
Huaizu Jiang
\qquad
Subhransu Maji\\
University of Massachusetts, Amherst\\
{\tt\small \{jcsu,chenyun,hzjiang,smaji\}@cs.umass.edu}\\
}

\maketitle
\thispagestyle{empty}
\begin{abstract}
We present a framework for learning to describe fine-grained visual differences between instances using \emph{attribute phrases}. 
Attribute phrases capture distinguishing aspects of an object (\eg, ``propeller on the nose" or ``door near the wing" for airplanes) in a compositional manner. 
Instances within a category can be described by a set of these phrases and collectively they span the space of semantic attributes for a category. We collect a large dataset of such phrases by asking annotators to describe several visual differences between a pair of instances within a category.
We then learn to describe and ground these phrases to images in the context of a
\emph{reference game} between a speaker and a listener. The goal of a speaker is to describe attributes of an image that allows the listener to correctly identify it within a pair.
Data collected in a pairwise manner improves the ability of the speaker to generate, and the ability of the listener to interpret visual descriptions.
Moreover, due to the compositionality of attribute phrases, the trained listeners can interpret descriptions not seen during training for image retrieval, and the speakers can generate attribute-based explanations for differences between previously unseen categories.
We also show that embedding an image into the semantic space of attribute phrases derived from listeners offers 20\% improvement in accuracy over existing attribute-based representations on the FGVC-aircraft dataset.
\end{abstract}
\vspace{-0.3in}
\section{Introduction}
\vspace{-0.1in}
\label{s:intro}
Attribute-based representations have been used for describing
instances within a basic-level category as they often share a set of high-level properties. 
These attributes serve as basis for 
human-centric tasks such as retrieval and
categorization~\cite{wah2014similarity,kovashka2012whittlesearch,parikh11relative},
and for generalization to new categories based on a description of their attributes~\cite{farhadi2009describing,farhadi2010attribute,sadeghi2011recognition,lampert2014attribute}.
However, most prior work has relied on a fixed set of attributes designed by experts.
This limits their scalability to new domains since collecting expert
annotations are expensive, and results in models that are less robust
to noisy open-ended descriptions provided by a
non-expert user.

Instead of discrete attributes this work investigates the use
of \emph{attribute phrases} for describing instances. Attribute
phrases are short sentences that describe a unique semantic visual
property of an object (\eg, ``red and white color", ``wing near the
top"). 
Like captions, they can describe properties in a compositional manner,
but are typically shorter and only capture a single aspect. Like
attributes, they are modular, and can be combined in different ways to
describe instances within a category. 
Their compositionality allows the expression of large number of
properties in a compact manner. 
For example, colors of objects, or their parts, can be expressed by
combining color terms (\eg., ``red
and white", ``green and blue", \etc). 
A collection of these phrases constitutes the semantic space of
describable attributes and can be
used as a basis for communication between a human and computer for
various tasks.

\begin{figure}
\begin{center}
   \includegraphics[width=\linewidth]{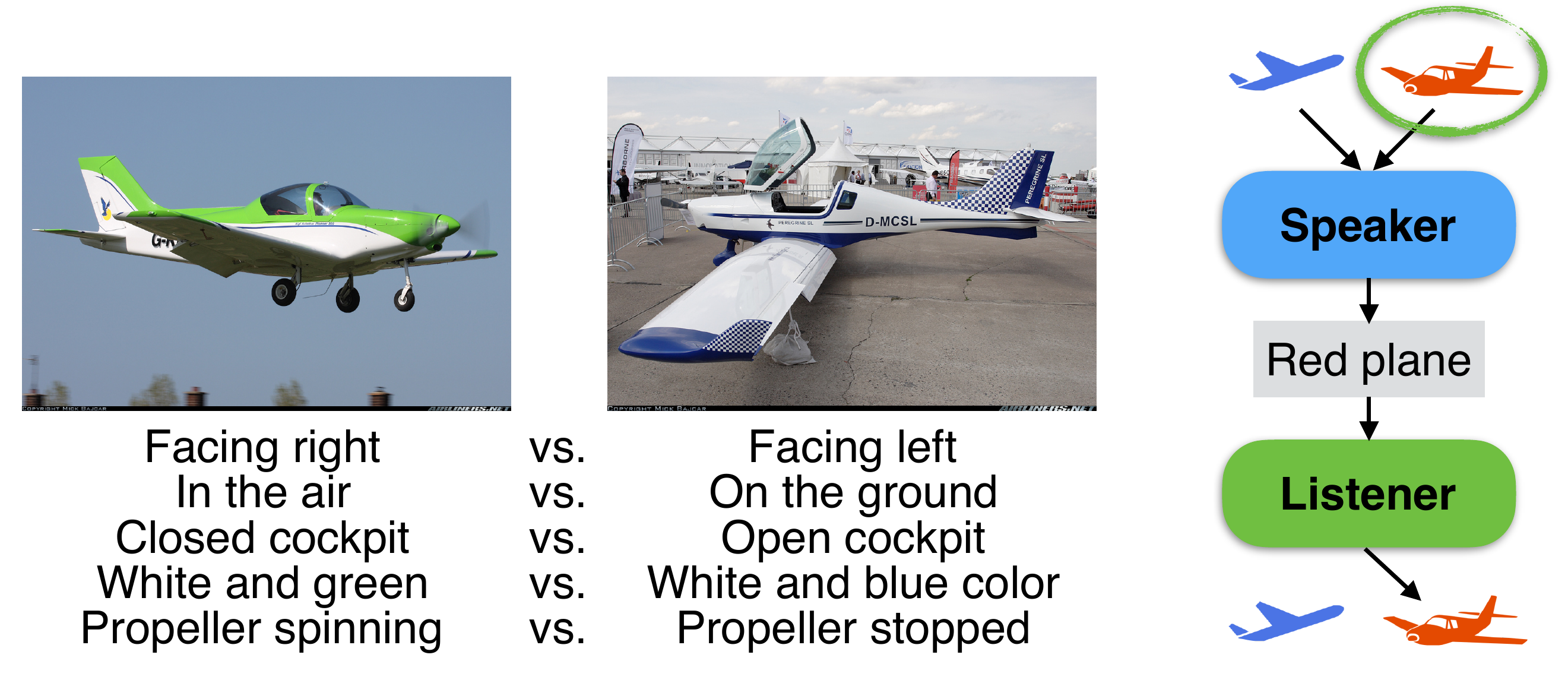}\\
\end{center}
   \caption{\label{fig:example} \emph{Left:} Each annotation in our
   dataset consists of five pairs of attribute phrases. \emph{Right:}
   A \emph{reference game} played between a speaker who describes an
   attribute of an image within a pair and a listener whose goal is to
   pick the right one.}
\vspace{-0.1in}
\end{figure}

We begin by collecting a dataset of attribute phrases by asking annotators to describe five visual differences between random pairs of airplanes from the OID airplane dataset~\cite{OID}.
Each difference is of the form ``P$_1$ vs.~P$_2$" with phrases P$_1$ and P$_2$ corresponding to the properties of the left and right image respectively (Figure~\ref{fig:example}). 
By collecting multiple properties at a time we obtain a diverse set of
describable attributes.
Moreover, phrases collected in a contrastive manner reveal attributes
that are better suited for fine-grained discrimination. 
The two phrases in a comparison describe the same underlying attribute
(\eg, \emph{round nose} and \emph{pointy nose} both describe the
shape), and reflect an axis of comparison in the underlying semantic
space. 
We then analyze the ability of automatic methods to generate these
attribute phrases using the collected dataset.
In particular we learn to generate descriptions and ground them in images in the context of a \emph{reference game} (RG) between a \emph{speaker} S and a \emph{listener} L~(Figure~\ref{fig:example}).
S is provided with a pair of images \{I$_1$, I$_2$\} and produces a visual difference of the form P$_1$ (or ``P$_1$ vs.~P$_2$"). L's goal is to identify which of the two images corresponds to P$_1$.
Reference games have been widely used to collect datasets describing
objects within a scene. 
This work employs the framework to generate and reason about compositional
language-based attributes for fine-grained visual categorization.

Our experiments show that a speaker trained to describe visual differences displays remarkable pragmatic behavior allowing a neural listener to rank the correct image with \textbf{91.4\%} \emph{top-5} accuracy in the RG compared with \textbf{80.6\%} of a speaker trained to generate captions non-contrastively. 
We also investigate a family of \emph{pragmatic speakers} who generate descriptions by jointly reasoning about the listener's ability to interpret them, based on the work of Andreas and Klein~\cite{Andreas2016}. 
Contrastively trained pragmatic speakers offer significant benefits (on average \textbf{7\%} higher \emph{top-5} accuracy in RG across listeners) over simple pragmatic speakers. The resulting speakers can be used to generate attribute-based explanations for differences between two categories. Moreover, given a set of attribute phrases, the score of an image with respect to each phrase according to a listener provides a natural embedding of the image into the space of semantic attributes.
On the task of image classification on the \emph{FGVC aircraft dataset}~\cite{maji2013fine} this representation outperforms existing attribute-based representations by \textbf{20\%} accuracy.

In summary, we show that reasoning about attribute phrases via reference games is a practical way of discovering and grounding describable attributes for fine-grained categories. We validate our approach on a dataset of 6,286 images with 9,400 pairs for a total of 47,000 phrases (Section~\ref{s:method}). We systematically evaluate various speakers and listeners using the RG and human studies (Section~\ref{s:listener}-\ref{s:speaker}), investigate the effectiveness of attribute phrases on various recognition tasks (Section~\ref{s:fgvc}-\ref{s:k2k}), and 
conclude in Section~\ref{s:conclusion}.

\vspace{-0.1in}
\section{Related Work}
\vspace{-0.1in}
\label{s:related}
\paragraph{Attribute-based representations.} Attributes have been widely used in the computer vision as an intermediate, interpretable representation for high-level recognition. They often represent properties that can be shared across categories, \eg, both a car and bicycle have wheels, or within a subordinate category, \eg, birds can be described by the shape of their beak. Due to their semantic nature they have been used for learning interpretable classifiers~\cite{farhadi2010attribute,farhadi2009describing}, attribute-based retrieval systems~\cite{chen2012describing}, as high-level priors for unseen categories for zero-shot learning~\cite{lampert2014attribute,jayaraman2014zero}, and as a means for communication in an interactive recognition system~\cite{kovashka2012whittlesearch}. 

A different line of work has explored the question of
discovering task-specific
attributes. Berg~\etal~\cite{berg2010automatic} discover attributes by
mining frequent n-grams in captions. 
Parikh and Grauman~\cite{parikh2011interactively} ask annotators to name directions that maximally separate the data according to some underlying features. Other approaches~\cite{sadeghi2015viske,izadinia2015segment,akata2015evaluation} have mined phrases from online text repositories to discover commonsense knowledge about properties of categories (\eg, cars have doors).
For a detailed description of the above methods see this recent survey~\cite{maji2017taxonomy}.

The interface for collecting attribute phrases is based on our earlier work~\cite{maji2012discovering}, which showed that annotations collected in a pairwise manner could be analyzed to discover a lexicon of parts and attributes. 
This work extends the prior work in a several ways. We (a) consider the complete problem of generating and interpreting attribute phrases on a significantly larger dataset, (b) systematically evaluate speaker and listener models on the data, and (c) show their utility in various recognition tasks.

\vspace{-0.1in}
\paragraph{Referring expression comprehension and generation.}
Modern captioning
systems~\cite{kiros2014unifying,donahue2015long,Vinyals2016} produce
descriptions by using encoder-decoder architectures, typically
consisting of a convolutional network for encoding an image and a
recurrent network for decoding a sentence. A criticism of these tasks
is that captions in existing datasets (\eg, MS COCO
dataset~\cite{lin2014microsoft}) can be generated by identifying the
dominant categories and relying on a language model. 
State-of-the-art systems are often matched by simple nearest-neighbor retrieval
approaches~\cite{devlin2015exploring,balanced_binary_vqa}. Visual
question-answering systems~\cite{VQA} face a similar issue that most
questions can be answered by relying on common-sense knowledge (\eg,
the sky is often blue). 
Some recent attempts have been made to address these
issues~\cite{johnson2017clevr}.

Tasks where annotators are asked
to describe an object in an image such that another can correctly
identify it provides a way
to collect context-sensitive
captions~\cite{kazemzadeh2014referitgame}. 
These tasks have been widely
studied in the linguistics community in an area called pragmatics (see
Grice's maxims~\cite{grice1975logic}). 
Much prior work in computer vision has focused on generating referring expressions to
distinguish an object within an
image~\cite{mitchell2013generating,kazemzadeh2014referitgame,mao2016generation,hu2016natural,nagaraja2016modeling,yu2016modeling,yu2017joint,luo2017comprehension}.
More recently, referring expression generation have been extended to interactive dialogue systems~\cite{das2017visual,vries2017GuessWhat}.
In contrast, our work aims to collect and generate referring expressions for
fine-grained discrimination between instances.

For the task of fine-grained recognition, the work of
Reed~\etal~\cite{Reed2016} is the most related to ours. They ask
annotators on Amazon Mechanical Turk to describe properties of birds
and flowers, and use the data to train models of images and
text. They show the utility of such models for zero-shot recognition
where a description of a novel category is provided as supervision,
and for text-based image retrieval.
Another recent work~\cite{vedantam2017context} showed that referring
expressions for images within a set can be generated simply by
enforcing separation of image probabilities during decoding using beam
search.
However, their model was trained on context agnostic captions.  
Our work takes a different approach. First, we collect attribute
phrases in a contrastive manner that encourages pragmatic behavior
among annotators. Second, we ask annotators to provide multiple
attribute descriptions, which as we described earlier, allows modular
reuse across instances, and serves as an intermediate representation
for various tasks. Attribute phrases capture the spectrum between
basic attributes and detailed captions. Like ``visual
phrases"~\cite{sadeghi2011recognition} they capture frequently
occurring relations between basic attributes.

\vspace{-0.1in}
\section{Method}
\vspace{-0.1in}
\label{s:method}
The framework used to collect our dataset is described in Section~\ref{s:dataset}. Various speaker and listener models are described in Section~\ref{s:speaker-model} and Section~\ref{s:listener-model} respectively.

\vspace{-0.1in}
\subsection{A dataset of attribute phrases}
\vspace{-0.1in}
\label{s:dataset}
We rely on human annotators to discover the space of descriptive attributes. 
Our annotations are collected on images from the OID aircraft dataset~\cite{OID}.
The annotations are organized into 4700 image pairs (1851 images) in training set, 2350 pairs (1730 images) in validation set, and 2350 pairs (2705 images) in test set. Each pair is chosen by picking two different images uniformly at random within the provided split in the OID aircraft dataset.

\begin{figure}
\begin{center}
   \includegraphics[width=\linewidth]{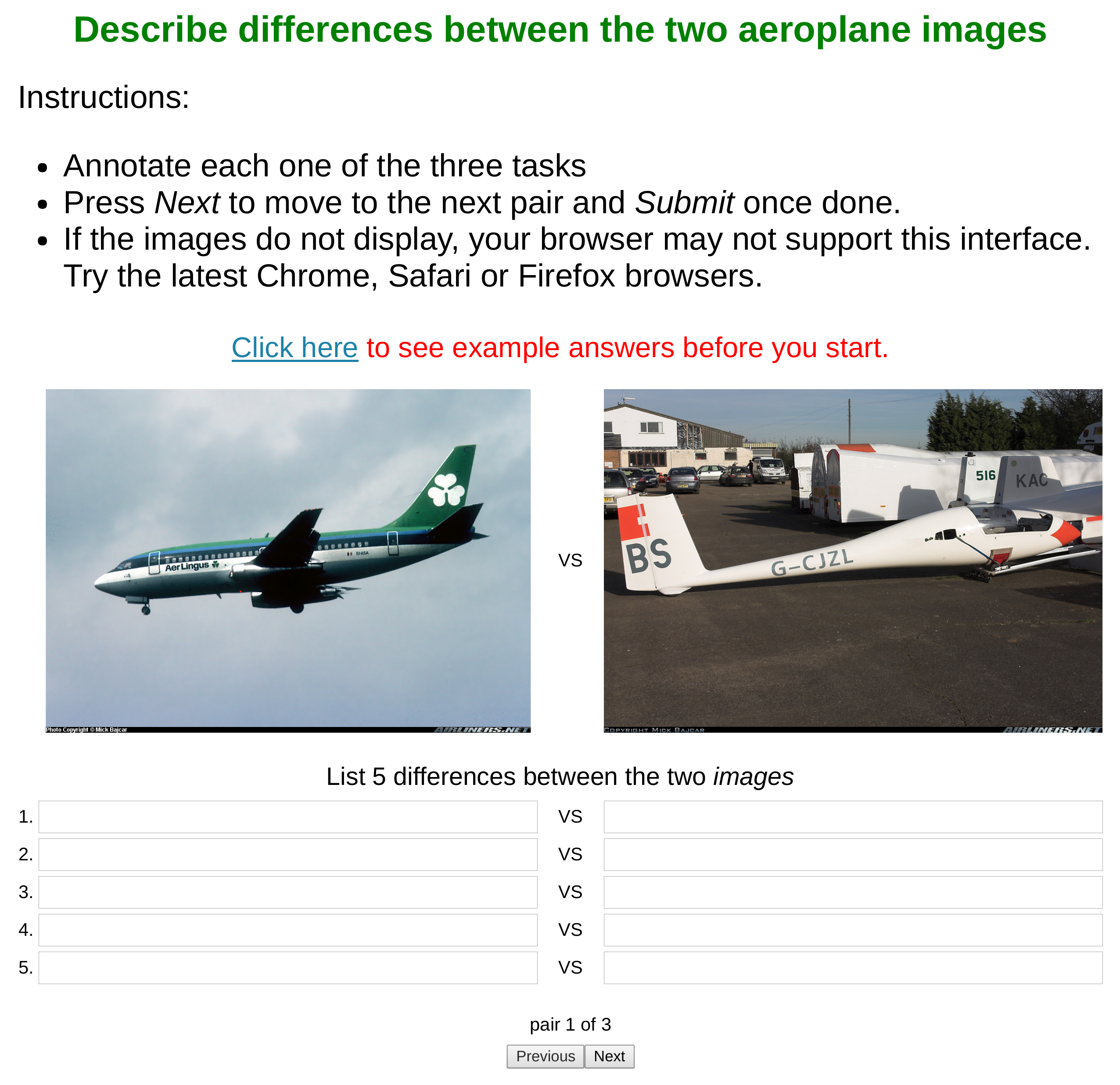}
\end{center}
   \caption{The annotation interface used to collect five different \emph{attribute phrase} pairs adapted from~\cite{maji2012discovering}. Amazon mechanical turkers were paid \$0.12 for annotating three pairs.}
\label{fig:interface}
\vspace{-0.1in}
\end{figure}

Annotators from Amazon Mechanical Turk are asked to describe five properties in the form ``P$_1$ vs.~P$_2$", each corresponding to a different aspect of the objects in the left and the right image respectively.
We also provide some examples as guidance to the annotators. 
The interface shown in Figure~\ref{fig:interface} is lightweight and allows rapid deployment compared to existing approaches for collecting attribute annotations where an expert decides the set and semantics of attributes ahead of time.
However, the resulting annotations are noisier and reflect the diversity of open-ended language-based descriptions.
A second pass over the data is done to check for consistency, after which  
about 15\% of the description pairs were discarded.

Figure~\ref{fig:example} shows an example of our dataset (more examples are in the supplementary material).
Annotations describe the shapes of parts (nose, wings and tail), relative sizes, orientation, colors, types of engines, \etc.
Most descriptions are short with an average length of 2.4 words on each side, although about 4.3\% of them have more than 4 words.
These are qualitatively different from image captions which are typically longer and more grammatical. However, each annotation provides five different attribute pairs.

The OID dataset also comes with a set of expert-designed attributes.
A comparison with OID attributes shows that attribute phrases capture novel properties that describe the relative arrangement of parts (\eg, ``door above the wing'', ``wing on top''), color combinations, relative sizes, shape, and number of parts (\eg, ``big nose'', ``more windows'', \etc) 
Section~\ref{s:embedding} shows a visualization of the space of attribute phrases. 
Section~\ref{s:fgvc} provides a direct comparison of OID attributes and those derived from our data on the task of FGVC-aircraft variant classification~\cite{maji2013fine}.

\subsection{Speaker models}\label{s:speaker-model}
\vspace{-0.1in}
A speaker maps visual inputs to attribute phrases. We consider two speakers; a \emph{simple speaker} (SS) that takes a single image as input and produces a description, and a \emph{discerning speaker} (DS) that takes two images as input and produces a single (or a pair of) description(s).

Both our speaker models are based on the show-and-tell model~\cite{Vinyals2016} developed for image captioning. Images are encoded using a convolutional network and decoded into a sentence using a recurrent network over words. 
We use one-hot encoding for 730 words with frequency greater than $5$ in the training set.
We consider \texttt{fc7} layer outputs of the VGG-16
network~\cite{simonyan2014very} plus two fully-connected layers with
ReLU units~\cite{nair2010rectified} on top as the image feature, and a
LSTM model~\cite{hochreiter1997long} with 2048 hidden units to
generate the sentences.  The image feature is fed into the LSTM not
only as the initial input, but also in each state input together with
word embeddings.
This led to an improved speaker in our
experiments. 
For the \emph{discerning speaker}, we concatenate two image features
as input to the LSTM. At test time we apply beam search with beam size
$10$ and get $10$ output descriptions from each image (pair). Although
the \emph{discerning speaker} is trained to generate phrase pairs, we
can simply take the first (or second) half of the pair and evaluate it
in the same way as a \emph{simple speaker}.

We also consider a \emph{pragmatic speaker} that generates
contrastive captions by reasoning about the listener's ability to pick
the correct image based on the description. 
Andreas and Klein~\cite{Andreas2016} proposed a simple strategy to do
so by reranking descriptions of an image based on a weighted
combination of (a) \emph{fluency} -- the score assigned by the
speaker, and (b) \emph{accuracy} -- the score assigned by the listener
on the referred image. Various pragmatic speakers are possible based
on the choice of speakers and listeners.
The details are described in Section~\ref{s:speaker}.

\emph{Optimization details:} Our implementation is based on
Tensorflow~\cite{tensorflow}. The descriptions are truncated at length
$14$ when training the LSTM. The VGG-16 network is initialized with
weights pre-trained on ImageNet
dataset~\cite{krizhevsky2012imagenet}. We first fix the VGG-16 weights
and train the rest of the network, using Adam
optimizer~\cite{kingma2014adam} with initial learning rate $0.001$,
$\beta_1= 0.7$, $\beta_2 = 0.999$ and $\epsilon = 1.0 \times 10
^{-8}$. We have batch normalization~\cite{ioffe2015batch} in fully connected layers after
VGG-16, and drop out with rate $0.7$ in LSTM. We use batch size $64$
for $40000$ steps ($\sim$28 epochs). Second, we fine tune the whole
network with initial learning rate modified to $5\times10^{-6}$, batch
size $32$ for another $40000$ steps.


\subsection{Listener models}\label{s:listener-model}
\vspace{-0.1in}
A listener interprets a single (or a pair of) attribute phrase(s), and
picks an image within a pair by measuring the similarity between the
phrase(s) and images in a common embedded space. Once again we
consider two listeners: a \emph{simple listener} (SL) that interprets
a single phrase, and a \emph{discerning listener} (DL) that interprets
a phrase pair. The \emph{simple listener} models the score of the
image I$_1$ within a pair (I$_1$, I$_2$) for a phrase $P$ as: 
\[
p(\text{I}_1|P) =
\sigma(\phi(\text{I}_1)^T\theta(\text{P}),
\phi(\text{I}_2)^T\theta(\text{P})).
\]
 Here $\phi$ and $\theta$ are
embeddings of the image and the phrase respectively, and $\sigma$ is
the \texttt{softmax} function $\sigma(x,y) =
\exp(x)/(\exp(x)+\exp(y))$. Similarly, a \emph{discerning listener}
models the score of an image by comparing it with an embedding of the
phrase pair $\theta([\text{P}_1 \text{ vs. } \text{P}_2])$. A simple
way to construct a discerning listener from a simple listener is by
averaging the predictions from the left and right phrases, \ie,
$p(\text{I}|[\text{P}_1 \text{ vs. } \text{P}_2]) =
\left(p(\text{I}|\text{P}_1) + p(\text{I}|\text{P}_2)\right)/2$.

We follow the setup of the speaker to embed phrases and use the final state of a LSTM with 1024 hidden nodes as the phrase embedding. The vocabulary of words is kept identical. For image features, once again we use the \texttt{fc7} layer of the VGG-16 network and add a fully-connected layer with 1024 units and ReLU activation. The parameters are learned by minimizing the cross-entropy loss. 

We also evaluate two variants of the \emph{simple listener}, SL$_r$ and SL, based on whether it is trained on non-contrastive data (I$_1$, I$_2$, P$_1$) where I$_2$ is a \emph{random image} within the training set, or the contrastive data where I$_2$ is the other image in the annotation pair. 

\emph{Optimization details:} We first fix the VGG-16 network and use Adam optimizer with initial learning rate $=0.001$, $\beta_1=0.7$, batch size $=32$ for $2000$ steps ($4000$ steps for SL$_r$ model), then fine-tune the entire model with a learning rate $1\times10^{-5}$ for another $7000$-$10000$ steps. 

\paragraph{Human listener.} We also consider human annotators to perform the task of the listener in the RG. For each generated phrase that describes one image out of an image pair, we let three users to pick which image out of the pair the phrase is referring to. However, unlike (most) human speakers, neural speakers can produce irrelevant descriptions. Thus, in addition to the choice of left and right image, users have the option to say \emph{``not sure''} when the description is ambiguous. If two or more users out of three picked the same image, we say the human listener is certain about the choice, otherwise we say the human listener is uncertain. The interface is shown in the supplementary material.

\section{Results}
\label{s:results}
\vspace{-0.1in}
We evaluate various listeners and speakers on the dataset we collected in terms of their accuracy in the RG in Section~\ref{s:listener} and Section~\ref{s:speaker} respectively. We then evaluate their effectiveness on a fine-grained classification task in Section~\ref{s:fgvc}, visualize the space of attribute phrases discovered from the data in Section~\ref{s:embedding}, for text-based image retrieval in Section~\ref{s:retrieval}, and for generating visual explanations for differences between categories in Section~\ref{s:k2k}.

\subsection{Evaluating listeners}\label{s:listener}
We first evaluate various listeners on human-generated phrases. 
For simple listeners, each annotation provides ten different reference tasks (I$_1$, I$_2$, P) $\rightarrow$ \{0,1\} corresponding to five different left and right attribute phrases. 
Each task is evaluated independently and accuracy is measured as the fraction of correct references made by the listener. 
Similarly, discerning listeners are evaluated by replacing P with
``P$_1$  vs. P$_2$" or ``P$_2$  vs. P$_1$".

\paragraph{Accuracy using human speakers.} The results are shown in Table~\ref{tab:listener_result}. Training on contrastive data improves the accuracy of the simple listener slightly from 84.2\%~(SL$_r$) to 86.3\%~(SL) on the test set.
Discerning listeners see both phrases at once and naturally perform better. There is almost no difference between a discerning listener that combines two simple listeners by averaging their predictions (2$\times$SL), and one that interprets the two phrases at once (DL).
The results indicate that on our dataset the listener's task is relatively easy and contrastive data does not provide any significant benefits. As a reference the accuracy of a human listener is close to 100\% on human-generated phrases. 

\begin{table}[h!]
\centering
\begin{tabular}{c|c|c|c|c}
Input & Speaker & Listener & Val & Test\\ 
\Xhline{3\arrayrulewidth}
\multirow{2}{*}{P$_1$} & \multirow{2}{*}{Human} & SL$_r$ 	& 82.7 & 84.2 \\
& & SL		& 85.3 & 86.3 \\
\hline
\multirow{2}{*}{P$_1$ vs. P$_2$} & \multirow{2}{*}{Human} & DL & 88.7 & 88.9 \\ 
& & 2$\times$SL &89.6 & 89.3 \\
\end{tabular}
\caption{Accuracy (\%) of various listeners in the RG using attribute phrases provided by a human speaker.}
\label{tab:listener_result}
\vspace{-0.15in}
\end{table}

\vspace{-0.08in}
\paragraph{Are the top attributes more salient?} As annotators
are asked to describe five different attributes they might pick ones
that are more salient first. 
We evaluate this hypothesis by measuring
the accuracy of the listener (SL) on phrases as a function of the
position of the annotation in the interface ranging from one for the top
attribute to five for the last one.
The results are shown in Table~\ref{tab:salient_result}. 
The accuracy decreases
monotonically from one to five suggesting that the first attribute
phrase is easier for the listener to discriminate. 
We are uncertain if this is because the attributes near the top are
more discriminative, or because the listener is better at interpreting
these as they are likely to be more frequent in the training
data. Nevertheless, attribute saliency is a signal we did not model
explicitly and may be used to train better speakers and listeners
(\eg, see Turakhia and Parikh~\cite{turakhia2013attribute}).

\begin{table}[h!]
\centering
\begin{tabular}{c|ccccc}
    & 1 & 2 & 3 & 4 & 5 \\
\Xhline{3\arrayrulewidth}
Val & 91.3 & 86.6 & 84.1 & 82.5 & 82.3 \\
Test  & 92.3 & 87.4 & 85.9 & 84.0 & 81.6 \\
\end{tabular}
\caption{Accuracy (\%) of the simple listener (SL) on RG using
human-generated attribute phrases at positions one through five across
the validation and test set. The accuracy decreases monotonically from
one to five suggesting that the top attribute phrases are easier to
discriminate.}
\vspace{-0.15in}
\label{tab:salient_result}
\end{table}


\subsection{Evaluating speakers}\label{s:speaker}
\vspace{-0.02in}
We use simple listeners, SL and SL$_r$, and the human listener to
evaluate speakers. 
As described in Section~\ref{s:speaker-model} we use beam search to generate $10$ descriptions for each image pair and evaluate them individually using various listeners.
The discerning speaker generates phrase pairs but we simply take the
first and second half separated by ``vs.", a special word in the
vocabulary, and evaluate it using a simple listener (that sees only
one phrase). If the word ``vs." is missing in the generated output we
simply consider the entire sentence as the P$_1$.
Only 1 out of 23500 phrase pairs did not contain the ``vs.'' token.

\begin{figure}[h!]
\begin{center}
\includegraphics[clip, width=\linewidth]{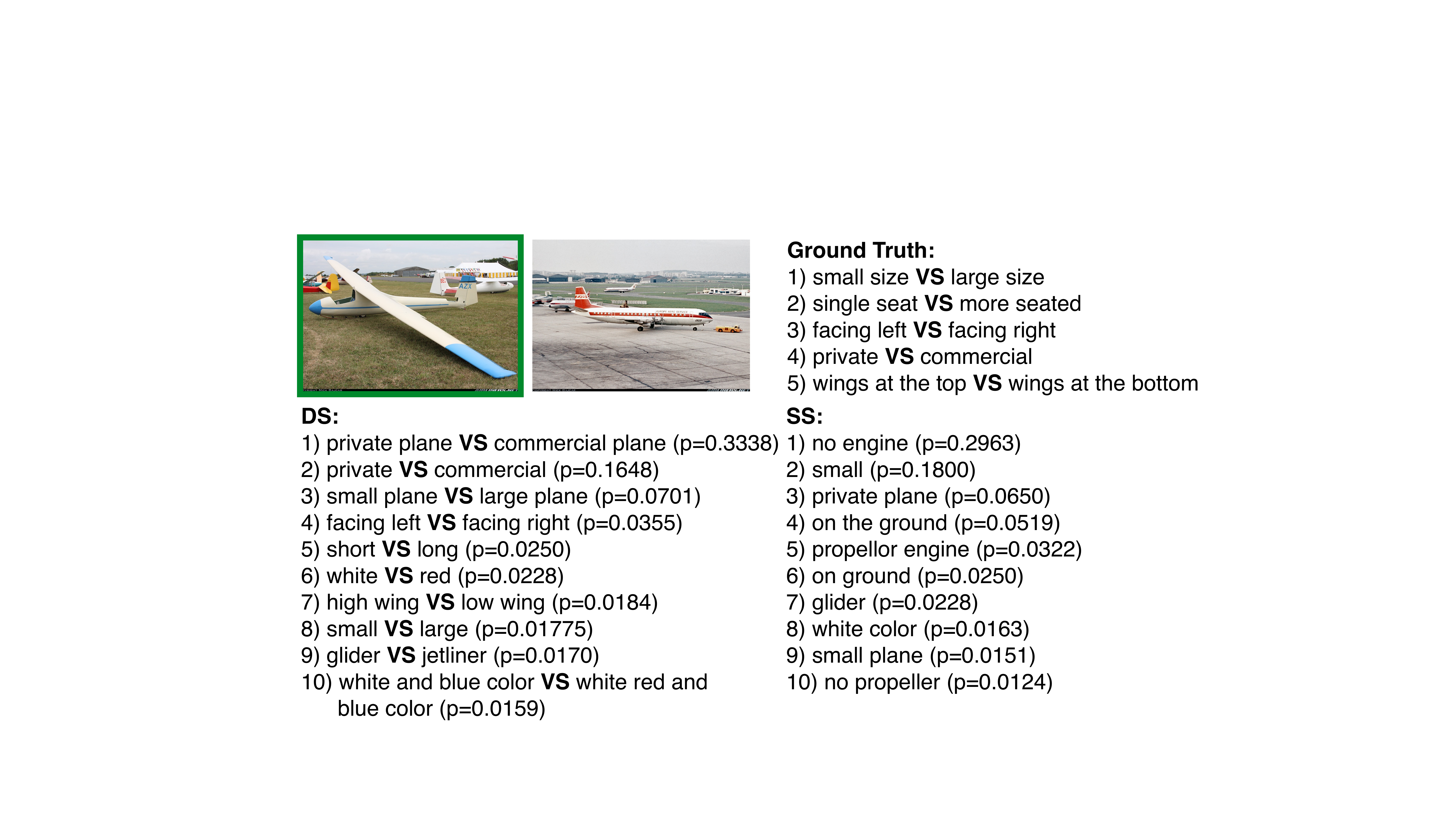}
\end{center}
\caption{Example output of simple speaker SS and discerning speaker DS. Simple speaker takes the left image in the green box as input, while the discerning speaker takes both images as input. In brackets are the probabilities according to the speaker.}
\label{fig:speaker}
\vspace{-0.2in}
\end{figure}

For evaluation with humans we collect three independent annotations on
a subset of 100 image pairs (with 10 descriptions each) out of the
full test set. 
The listeners are considered to be correct when the
probability of the correct image is greater than half. 
For human listener, we report the accuracy of when there is a majority
agreement on the correct image, \ie, when two or more users picked the
correct image.
For direct comparison with the simple speaker models, we also report the human listener accuracy when they are allowed to guess. This is the sum of earlier accuracy, and half of the cases when there is no majority agreement.
Human annotators are uncertain when the generated descriptions are not
fluent or when they are not discriminative. 
Therefore, a better human accuracy reflects speaker quality
both in terms of fluency and discriminativeness.
Some examples of the
generated attribute phrases using various speakers are shown in Figure~\ref{fig:speaker}.

\paragraph{Accuracy of various speakers and listeners.} Results on the full test set (Test) and the human-evaluated subset (Test*) are shown in Table~\ref{tab:speaker_result}. The accuracy of discerning speaker exceeds that of simple speaker by more than $10\%$ no matter which listener to use.
This result suggests that data collected contrastively using our
annotation task allows direct training of speaker models that show
remarkable context-sensitive behavior.  Somewhat surprisingly we also
see that the simple listeners are more accurate than the human
listener when evaluated on descriptions generated by our speaker
models. This is because humans tend to be more cautious in the
reference game. For example, simple listeners will accept yellowish
grass being referred to as ``concrete'' compared to green grass, but
humans tend to view it as an unclear reference.

\begin{table}
\centering
\begin{tabular}{cccc|cc|c}
\Xhline{3\arrayrulewidth}
& 				& 		\multicolumn{5}{c}{\textbf{Accuracy (\%)}}\\ \cline{3-7}
& 				& 		\multicolumn{2}{c|}{SL$_r$}	& \multicolumn{2}{c|}{SL}    & Human \\ 
	 		& Top				& Test$^*$	& Test		& Test$^*$	& Test		&Test$^*$ \\ 
\Xhline{3\arrayrulewidth}
\multirow{3}{*}{SS}	& 1		& 84.0		& 79.8		& 83.0 		& 81.7		& 68.0 (77.0)\\
						& 5		& 80.0		& 79.2		& 78.0 		& 80.6		& 64.2 (74.1)\\
						& 10	& 78.0		& 78.9		& 76.6 		& 80.0		& 61.6 (72.4)\\ 						
\hline
\multirow{3}{*}{DS}	& 1		& 94.0		& 92.8		& 92.0 		& 92.8		& 82.0 (88.5)\\
						& 5		& 91.2		& 90.3		& 91.2 		& 91.4		& 80.2 (86.7)\\
						& 10	& 88.6		& 88.8		& 90.0 		& 90.5		& 77.9 (85.0)\\ 
\end{tabular}
\caption{Accuracy in the RG using different speakers and listeners. 
Test represents the full test set consisting of $2350$ image pairs. Test$^*$ represents a subset of $100$ test set image pairs for which we collected human listener results. 
For the human listener, we report the accuracy when there is a majority agreement, and accuracy with guessing (in brackets).
DS is significantly better at generating discriminative attribute
phrases than SS.}
\label{tab:speaker_result}
\vspace{-0.05in}
\end{table}


\paragraph{Does pragmatics help?} Given that our discerning speaker
can generate highly accurate contrastive descriptions, we investigate
if additional benefits can be achieved if the speaker jointly reasons
about the listener's ability to interpret the descriptions. We employ
the \emph{pragmatic speaker} model of Andreas and
Klein~\cite{Andreas2016} where a simple speaker generates descriptions
that are reranked by a simple listener using a weighted combination
of speaker and listener scores. In particular, we rerank the output
$10$ sentences from speakers by the probabilities from simple
listeners. 
We combine the listener probability $p_l$ and speaker beam-search
probability $p_s$ as $p = p_s^\lambda \cdot p_l^{(1-\lambda)}$, and
pick the optimal $\lambda$ on a validation set annotated by a human
listener. We found that the optimal $\lambda$ is close to $0$, so we
decided to use $p_l$ only for reranking on test set.

In Table~\ref{tab:pragmatic_result1}, we report the accuracy of top $k$ sentences ($k=1, 5, 7$) of the human listener and the results after reranking on the Test* set. 
When using the listener score from SL$_r$ the average accuracy of the
top five generated descriptions after reranking improves dramatically
from 64.2\% to 82.6\% for the simple speaker. The accuracy of the
discerning speaker also improves to 90\%. This suggests that better
pragmatics can be achieved if both the speaker and listener are
trained in a contrastive manner. Surprisingly the
contrastively-trained simple listener SL is less effective at
reranking than SL$_r$. We believe this is because the SL overfits on
the human speaker descriptions and is less effective when used with
neural speakers.

\begin{figure*}[h!]
\begin{center}
\includegraphics[clip, width=0.9\linewidth]{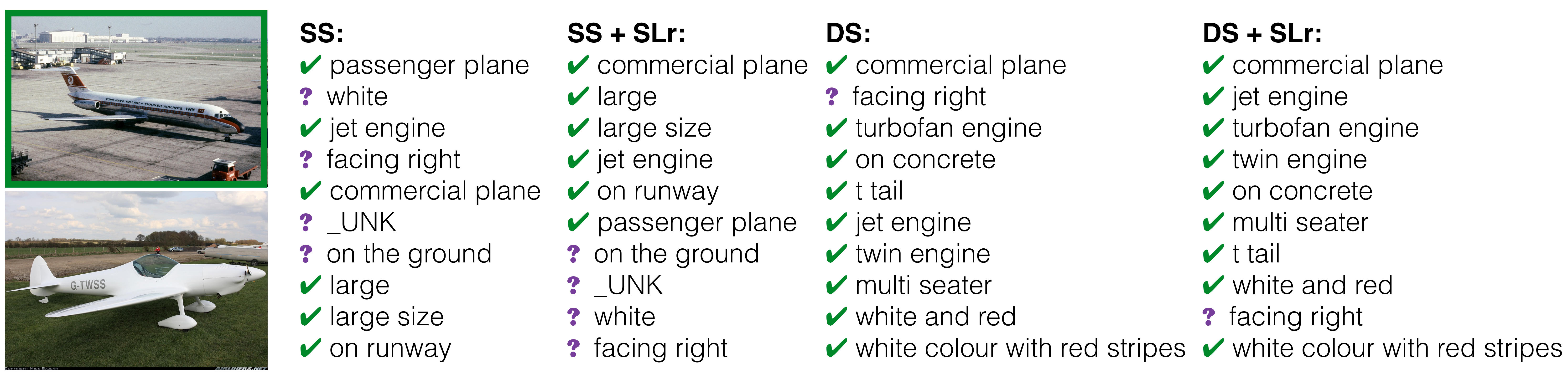}
\end{center}
\caption{An example output of various speakers. Given the image pair, we use SS and DS
				to generate descriptions of the top
				left image.
Outputs from SS and DS are listed in the order of probabilities from
				speaker beam search. Outputs of
				SS+SL$_r$ and DS+SL$_r$ are reranked
				by SL$_r$. 
Green checks mean human listener picks correct image with certain,
				while question marks mean human
				listener is uncertain which image is
				referred to.
The results indicate that DS is better than SS, and reranking using
				listeners improves the quality of top
				sentences.}

\label{fig:pds}
\vspace{-0.05in}
\end{figure*}

Figure~\ref{fig:pds} shows an example pair and the output of different
speakers. Simple speaker suffers from generating descriptions that are
true to the target image, but fail to differentiate two
images. Discerning speaker can mostly avoid this mistake. Reranking by
listeners can move better sentences to the top and improves the
quality of top sentences.

\begin{table}[t!]
\centering
\begin{tabular}{ccc|c|c}
\Xhline{3\arrayrulewidth}
 & & \multicolumn{3}{c}{\textbf{Human listener accuracy (\%)}} \\ \cline{3-5}
 & & \multicolumn{3}{c}{Reranker listener} \\ 
 			& Top 	& None		& SL$_r$		& SL \\ 
\Xhline{3\arrayrulewidth}
\multirow{3}{*}{SS}	&  1		& 68.0 (77.0)	& 94.0 (96.0)	& 87.0 (92.0) \\
				&  5		& 64.2 (74.1)	& 82.6 (88.3)	& 80.8 (87.1) \\
				&  7		& 63.1 (72.8)	& 74.3 (82.0)	& 74.3 (82.4) \\ \hline
\multirow{3}{*}{DS}	&  1		& 82.0 (88.5)	& 95.0 (96.5)	& 95.0 (97.0) \\
				&  5		& 80.2 (86.7)	& 90.0 (93.3)	& 88.6 (92.8) \\
				&  7		& 79.1 (85.6)	& 86.7 (91.5)	& 86.1 (91.1) \\ 
\end{tabular}
\caption{Accuracy of pragmatic speakers with human listeners on the
				Test* set.  After generating the
				descriptions by the speaker model
				(either SS or DS), we use the listener
				model (SL$_r$ or SL) to rerank
				them. We report the accuracy based on
				human listener from the user study. We
				report both the accuracy when there is
				majority agreement, and accuracy with
				guessing (in brackets). Pragmatic
				speakers are strictly better than
				non-pragmatic ones.}

\label{tab:pragmatic_result1}
\end{table}

%

\subsection{Fine-grained classification with attributes}\label{s:fgvc}
\vspace{-0.1in}
We compare the effectiveness of attribute phrases to existing attributes in the OID dataset on the task of fine-grained classification on the FGVC aircraft dataset~\cite{maji2013fine}.
The OID dataset is designed with attributes in mind and has long-tail distribution over aircraft variants with 2728 models, while the FGVC dataset is designed for fine-grained classification task with 100 variants each with 100 images.
Both datasets are based on the images from the \url{airliners.net} website and have a few overlapping images.
We exclude the 169 images from the FGVC test set that appear in the OID training+validation set in our evaluation.

There are 49 attributes in the OID dataset organized into 14
categories. We exclude three attributes -- two referring to the
airline label and model, most of which have only one training examples
per category, and another that is rare. We then trained linear
classifiers to predict each attribute using the \texttt{fc7} layer
feature of the VGG-16 network. Using the same features and trained
classifiers, we construct a 46 dimensional embedding of the FGVC
images into the space of OID attributes. 
The attribute classifiers based on the VGG-16 network features are fairly
accurate (66\% mean AP across attributes) and outperforms the Fisher
vector baseline included in the OID dataset paper.

For the attribute phrase embeddings, we first obtain the K most frequent ones in our training set. Given an image I, we compute the score $\phi(\text{I})^T\theta(\text{P})$ for each phrase P from a listener as the embedding.
For a fair comparison the image features are kept identical to the OID attribute classifiers. We also explore an \emph{opponent attribute space}, where instead of top phrases we consider the top phrase pairs. Phrase pairs represent an axis of comparison, \eg, ``small vs.~medium", or ``red and blue vs.~red and white", and are better suited for describing relative attributes.  We use the discerning listener for the embedding on the opponent attribute space.

\begin{figure}
\vspace{-0.2in}
\begin{center}
   \includegraphics[width=\linewidth]{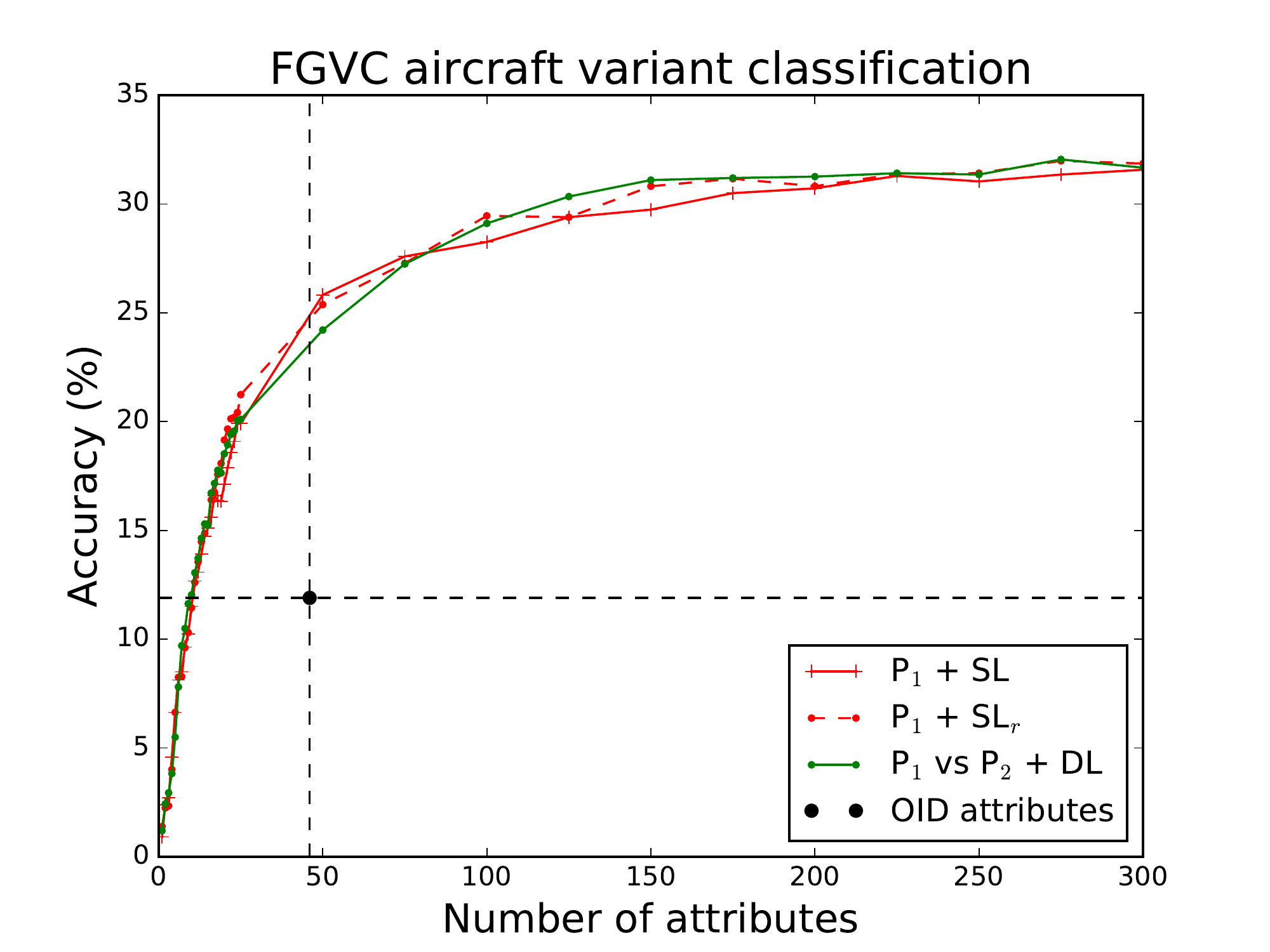}\\
\end{center}
   \caption{Classification accuracy on FGVC aircraft dataset using the 46 dimensional OID attributes and varying number of attribute phrases. See Section~\ref{s:fgvc} for details.}
   \label{fig:fgvc}
\vspace{-0.15in}
\end{figure}

Figure~\ref{fig:fgvc} shows a comparison of OID attributes and
attribute phrases for various listeners and number of attributes. 
For the same number of attributes as the OID dataset, attribute
phrases are \textbf{12\%} better. 
With 300 attributes the accuracy improves to \textbf{32\%},
about \textbf{20\%} better than OID. 
These results indicate that attribute phrases provide a better
coverage of the space of discriminative directions. The two simple
listeners perform equally well and the opponent attribute space does
not offer any additional benefits.

\subsection{Visualizing the space of descriptive
   attributes}\label{s:embedding}
\begin{figure}
\begin{center}
   \includegraphics[width=0.9\linewidth]{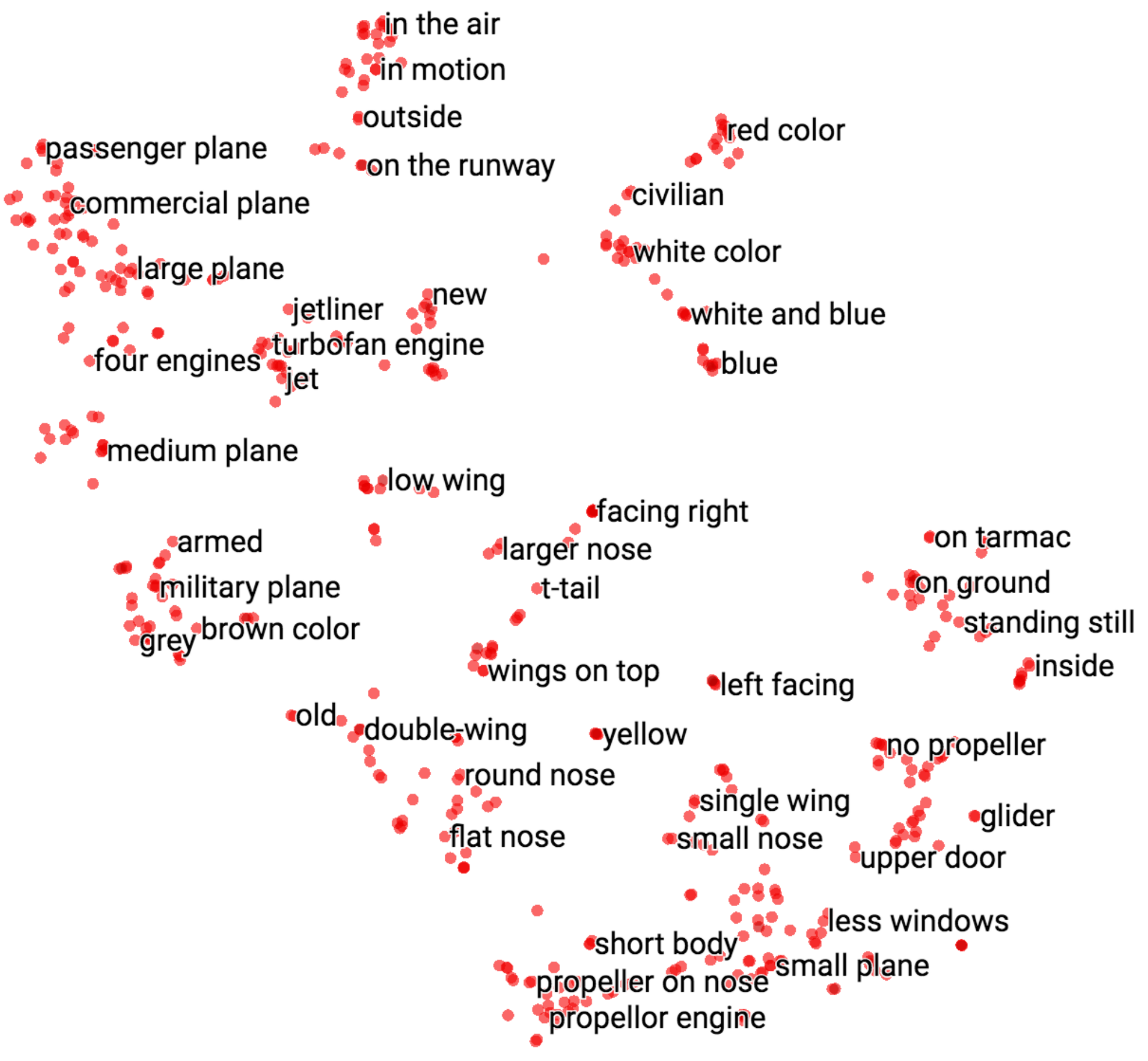}\\
\end{center}
   \caption{\label{fig:sent_embedding} Visualization of the $500$ most
   frequent descriptions. Each attribute is embedded into a $1024$
   dimensional space using the simple listener SL and projected into
   two dimensions using t-SNE~\cite{tSNE}. (\emph{Best viewed
   digitally with zoom.})}
\label{fig:embedding}
\end{figure}

\vspace{-0.05in}
We visualize the space of the 500 most frequent phrases in the training
set using the embedding of the simple listener model projected from
$1024$ dimensions to $2$ using t-SNE~\cite{tSNE} in
Figure~\ref{fig:embedding}. Various semantically related phrases are
clustered into groups. The cluster on the top right reflects color
combinations; Phrases such as ``less windows" and ``small plane"
are nearby (bottom right). Visualizations of the learned embeddings of
images $\phi(\text{I})$ and opponent attribute phrases
$\theta([\text{P}_1\text{ vs. } \text{P}_2])$ are provided in the
supplementary material.

\begin{figure*}
\begin{center}
   \includegraphics[width=0.825\linewidth]{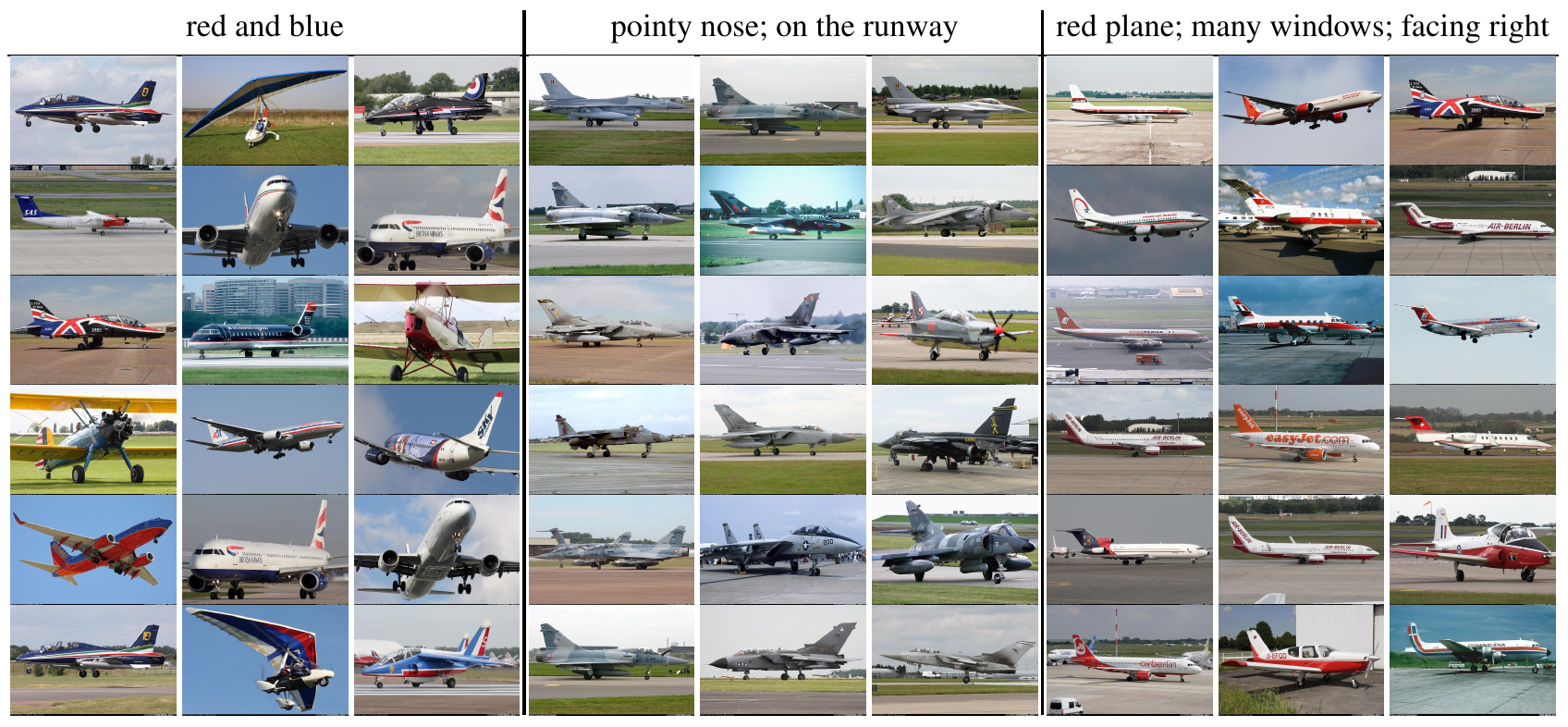}\\
\end{center}
   \caption{Top 18 images ranked by the listener for various attribute phrases as queries (shown on top). We rank the images by the scores from the simple listener on the concatenation of the attribute phrases. The images are ordered from top to bottom, left to right.}
\label{fig:img_retrieval}
\end{figure*}

\begin{figure*}[h!]
\begin{center}
   \includegraphics[width=\linewidth]{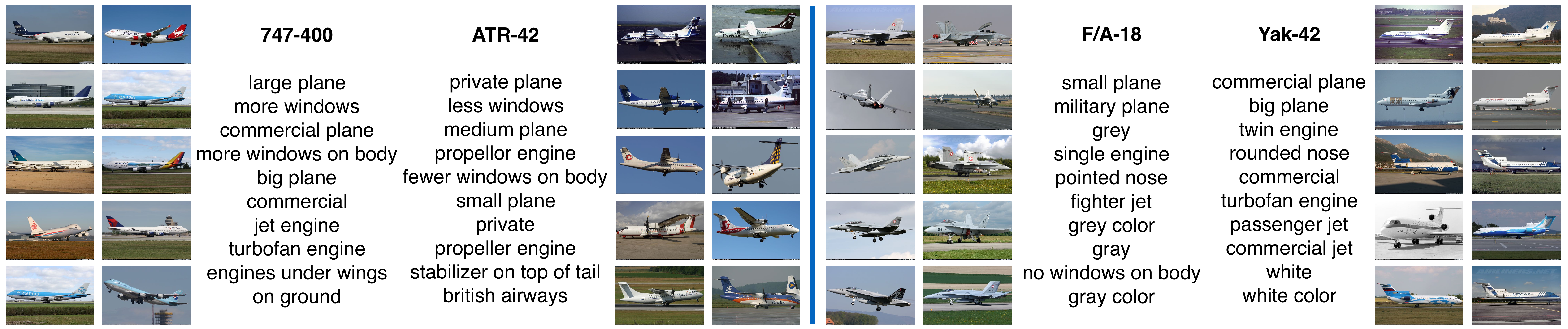}\\
\end{center}
   \caption{Top 10 discriminative attribute phrases for pairs of categories from FGVC aircraft dataset. Descriptions are generated by the discerning speaker for each pair of images in the first and second category. The phrases sorted by the occurrence frequency provides an attribute-based explanation of the visual difference between two categories.}
\label{fig:k2k}
\vspace{-0.15in}
\end{figure*}

\subsection{Image retrieval with descriptive attributes}\label{s:retrieval}
\vspace{-0.1in}
The listeners also allows us to retrieve an image given one or more attribute phrases. Given a phrase P we rank the images in the test set by the listener scores $\phi(\text{I})^T\theta(\text{P})$. Figure~\ref{fig:img_retrieval} shows some query phrases and the 18 most similar images retrieved from the test set. These results were obtained by simply concatenating all the query phrases to obtain a single phrase. More sophisticated schemes for combining scores from individual phrase predictions are likely to improve results~\cite{scheirer2012multi}. Our model can retrieve images with multiple attribute phrases well even though the composition of phrases does not appear in the training set. For example, ``red and blue" only shows five times in total of $47,000$ phrases in the training set, ``pointy nose" and ``on the runway" are never seen in a single phrase together. 

\subsection{Generating attribute explanations}\label{s:k2k}
\vspace{-0.1in}
The pairwise reasoning of a speaker can be extended to analyze an instance within a \emph{set} by aggregating speaker utterances across all pairs that include the target. 
Similarly one can describe differences between two sets by considering all pairs of instances across the two sets. 
We use this to generate attribute-based explanations for visual differences between two categories. We select two categories $A,B$ from FGVC aircraft dataset and randomly choose ten images from each category. 
For each image pair $(\text{I}_1 \in A, \text{I}_2 \in B)$, we generate ten phrase pairs using our discerning speaker. We then sort unique phrases primarily by their image frequency (number of images from target category described by the given description minus that from the opposite category), and when tied secondarily by their phrase frequency (number of occurrences of the phrase in target category minus that in the opposite category.)
The top ten attribute phrases for the two categories for an example pair of categories are shown in Figure~\ref{fig:k2k}.
The algorithm reveals several discriminative attributes between two such as ``engine under wings'' for 747-400, and ``stabilizer on top of tail" for ATR-42.

\section{Conclusion}
\label{s:conclusion}
\vspace{-0.15in}
We analyzed attribute phrases that emerge when annotators describe visual differences between instances within a subordinate category (airplanes), and showed that speakers and listeners trained on this data can be used for various human-centric tasks such as text-based retrieval and attribute-based explanations of visual differences between unseen categories.
Our experiments indicate that pragmatic speakers that combine listeners and speakers are effective on the reference game~\cite{Andreas2016}, and speakers trained on contrastive data offers significant additional benefits.
We also showed that attribute phrases are modular and can be used to embed images into an interpretable semantic space. 
The resulting attribute phrases are highly discriminative and outperform existing attributes on FGVC aircraft dataset on the fine-grained classification task.
\vspace{-0.05in}

{
\textbf{Acknowledgement: } This research was supported in part by the NSF grants 1617917 and 1661259, and a faculty gift from Facebook. The experiments were performed using equipment obtained under a grant from the Collaborative R\&D Fund 
managed by the Massachusetts Tech Collaborative and GPUs donated by NVIDIA.}

{\small
\bibliographystyle{ieee}
\bibliography{egbib,captioning_ref}
}

\end{document}